**Title:** Learning fast and agile quadrupedal locomotion over complex terrain

**Authors:**
Xu Chang[1,2]*, Zhitong Zhang[1,2], Honglei An[1], Hongxu Ma[1], Qing Wei[1]

**Affiliations:**

[1]Robotics Research Center, College of Intelligence Science and Technology, National University of Defense Technology, Changsha, Hunan, China.
[2]Co-first author.
*Corresponding author. Email: 457911161@qq.com

**Abstract:** In this paper, we propose a robust controller that achieves natural and stably fast locomotion on a real blind quadruped robot. With only proprioceptive information, the quadruped robot can move at a maximum speed of 10 times its body length, and has the ability to pass through various complex terrains. The controller is trained in the simulation environment by model-free reinforcement learning. In this paper, the proposed loose neighborhood control architecture not only guarantees the learning rate, but also obtains an action network that is easy to transfer to a real quadruped robot. Our research finds that there is a problem of data symmetry loss during training, which leads to unbalanced performance of the learned controller on the left-right symmetric quadruped robot structure, and proposes a mirror-world neural network to solve the performance problem. The learned controller composed of the mirror-world network can make the robot achieve excellent anti-disturbance ability. No specific human knowledge such as a foot trajectory generator are used in the training architecture. The learned controller can coordinate the robot's gait frequency and locomotion speed, and the locomotion pattern is more natural and reasonable than the artificially designed controller. Our controller has excellent anti-disturbance performance, and has good generalization ability to reach locomotion speeds it has never learned and traverse terrains it has never seen before.

**One-Sentence Summary:**
We classify reinforcement learning as a nonlinear solver from a mathematical point of view, discover the phenomenon of data symmetry loss during training phase, use the proposed mirror-world network to deal with the imbalanced performance, propose a loose neighborhood control architecture to make the simulation results easier to transfer to the physical robot, and our controller which is trained without any specific human knowledge can make the quadruped robot UnitreeA1 stably pass through various complex terrains, and the maximum locomotion speed can approach the hardware limit.

**Main Text:**

**INTRODUCTION**

Compared with wheeled mobile robots, legged robots have better flexibility and environment passability. Legged robots with natural locomotion and flexible operation capabilities can not only accompany human beings, but also work in place of humans in complex and dangerous environments. In order to cope with complex and changeable physical world scenarios, the robot needs a powerful locomotion controller as a foundation. Current locomotion controllers can be divided into two categories: manually designed controllers and learned controllers.

Manually designed controllers usually need to simplify complex robot dynamics to a form that humans can handle, and divide the control problem into several modules for processing (*1-7*). According to the information provided by the robot state estimation module, many elaborate state machines are set for different scenarios for switching (*8-14*). This "simplification-segmentation" processing method makes it difficult for quadruped robots to be applied to complex and changeable real-world environments, and the potential of robot hardware cannot be fully utilized. It is difficult for the artificially designed controller to coordinate the gait cycle, foot trajectories and balance of the supporting legs of the robot under different terrains and different motion speeds. Even a small increase in the locomotion speed is very expensive. To realize the highly dynamic and continuous complex motion of the robot, costs of time and labor are higher. Moreover, the designed motion patterns are not natural.

Recently, many practical reinforcement learning algorithms have been developed (*15-20*), controllers developed using model-free reinforcement learning have made some progress in locomotion control of legged robots (*21-26*). During the reinforcement learning training process, robots continuously interact with the simulation environment and collect data. According to the different rewards obtained at each time step, parameters of the controller are adjusted in the direction of maximizing the sum of the rewards. The training process can be performed in a professional and reliable simulation engine (*27-29*), and the entire nonlinear system control problem can be solved with full consideration of the complex dynamics of the robot. Some studies (*30*) propose teacher/student training strategies, and use proprioceptive information, privileged learning (*31*) and TRPO reinforcement learning algorithm (*16*) to realize that the quadruped robot ANYmal can pass through challenging terrains. These achievements on ANYmal allow the quadruped robot to go out of the laboratory and can provide a stable and reliable motion platform for real complex situations.

However, current learned controllers are mainly limited to low-speed, low-dynamic application scenarios. In contrast, the high-speed and high-dynamic motion of the robot puts forward higher requirements on the controller. As the speed of robot motion increases, the reality gap (*32, 33*) between the simulated environment and the real world will be magnified, and the magnified gap will have a significant impact on the robot locomotion performance. The significance of realizing a learned controller that can make the robot move fast is not only that the robot has the ability to run at high speed, but more importantly, it means that the highly dynamic continuous motion obtained by training in the simulation environment can be transferred to the real robot. That indicates the potential of robotic hardware can be fully exploited.

Our research focuses on how to train a high dynamic locomotion controller that is easy to transfer to a physical quadruped robot from the simulation environment, so that the robot can pass through various complex terrains, and give full play to the potential of the hardware to achieve the highest possible running speed. Our robot is UnitreeA1, whose effective body length (the distance between the front and rear thigh axes) is about 0.36m and the total weight is about 13.5kg. It has 12 rotary joints driven by DC motors, and the rated torque of the motor is $6.7\,\text{N}\cdot\text{m}$. Our proposed controller can make UnitreeA1 stably pass through various complex terrains by using only proprioception information such as IMU and joint sensors in the absence of exteroceptive sensors such as vision and laser radar. The trajectories of the feet and gait frequency of the quadruped robot are naturally coordinated with the locomotion speed. The maximum locomotion speed of the robot can reach 3.6m/s, which is close to the hardware limit.

To achieve our demonstrated results, there are three key concepts in addition to automatic curriculum learning (*34*). The first is to employ a different control architecture, which we call the loose neighborhood control architecture. Unlike previous work (*22, 30*), the loose neighborhood control architecture does not use any artificially designed gaits or foot trajectory generators. Since our control mode is not position control, there is no high-frequency joint torque that is easy to occur in joint position control, which makes the controller network easier to transfer to a physical robot.

The second key concept is the impact test training field. We find that in the process of robot simulation training, the phenomenon of unbalanced data samples will inevitably occur, resulting in that the performance of the learned controller cannot make the robot to maintain the same locomotion ability in symmetrical directions, especially when the robot is moving at high speed. By reasonably setting the parameters of the impact test training field, the robot's advantage motion area can be determined.

The third key concept is the mirror world neural network. After the robot's advantage motion area is determined through the impact test training ground, a mirror world network can be constructed to transform the robot's weakness area into an advantage area. Consequently, the anti-disturbance ability and locomotion ability of the robot are improved in an all-round way, and the robot can keep stable in high-speed motion and complex terrains.

# RESULTS

The experiments are divided into four parts in total, the first three parts are used to verify the performance of our controller in simulation and real world, and the fourth part is used to verify our proposed data symmetry loss phenomenon.

## Controller tracking performance

Our controller tracks the linear and angular speed commands. Fig. 1 shows the tracking performance of the controller in the simulation. Fig. 1A shows the results of linear velocity tracking. The robot runs straight from a stationary state along a fixed heading. The initial command speed is 1m/s, and then the command increases by 1m/s every second until it reaches 5.5m/s. On the premise that the front and rear legs do not collide, the maximum locomotion speed is greater than 4.7m/s. Fig. 1B shows the results of angular speeds tracking. The experiment is carried out in 8 groups, and each group is set with different angular speed commands. The robot starts from the origin and keeps the linear velocity command at 1m/s. After the heading rotates 360 degrees, the robot comes to a stop and records the consumption time $t$. The mean angular speed can be obtained as $2\pi/t$. The results show that the proposed controller has good tracking performance.

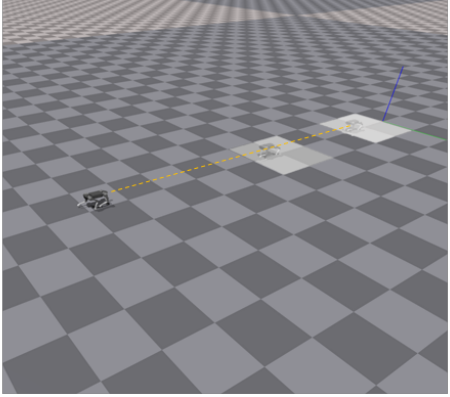
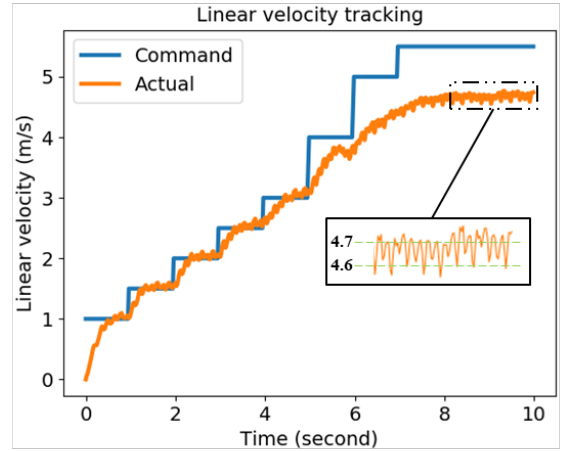
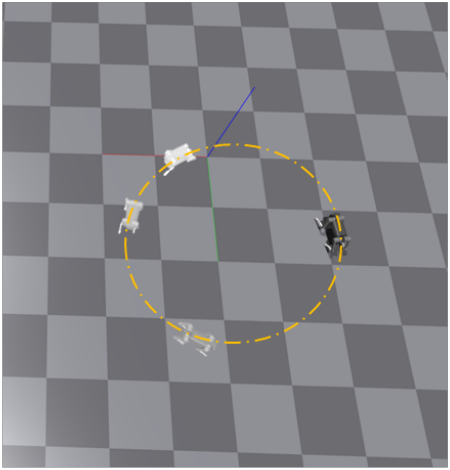
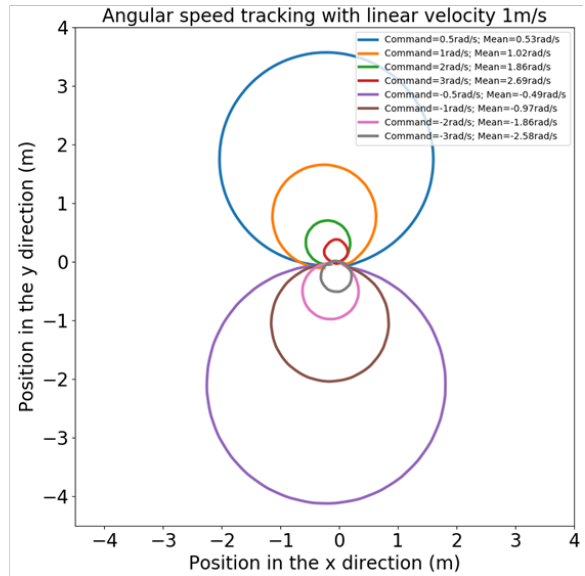

**Fig. 1. Linear velocity and angular speed tracking performance.** (**A**) Results of linear velocities tracking. (**B**) Results of angular speeds tracking.

**Outdoor environments**
Movie S1 shows the performance of the robot running at high speed on the artificial turf. The robot is in contact with floating granules of the artificial turf, and it is easier to cause sliding than normal ground. Locomotion on this ground mainly examines the high-speed stability of our controller. As the command speed increases, the gait frequency and foot trajectories of the robot change naturally with the body speed during the locomotion, as shown in Fig. 2. In the training phase, the range of linear velocity command is -2m/s to 2m/s. The learned controller has good motion generalization ability, and it can make the robot run at a maximum running speed of 3.6m/s which is a speed that has never been trained before and is close to the hardware limit (Motors cannot provide sufficient lasting torques at higher speed. Continuing to increase the speed will cause the motherboard to restart). Using our proposed mirror world network, the robot obtains a strong ability to stabilize. Even when the robot is close to falling, it can still restore its balance and continue to run.

Movie S2 shows that our controller has good adaptability and can pass through complex terrains such as mud, grass, and gravel areas that have never been seen before in training, as shown in Fig. 3A~3E. When passing through high grass with a similar height to it, the robot can maintain its balance even if its legs are entangled by grass or are stuck accidentally by a branch. The proposed controller can make the robot move continuously and stably in the urban outdoor environment with uneven ground and slopes ($< 20°$), and there is no single fall until the power is exhausted.

**A**. Low speed foot trajectories

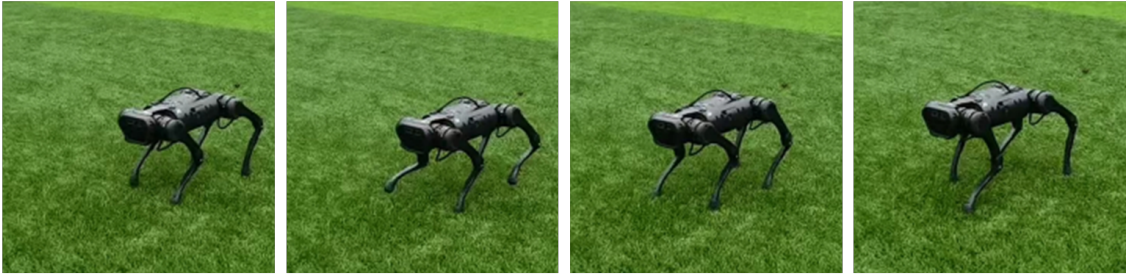

**B**. High speed foot trajectories

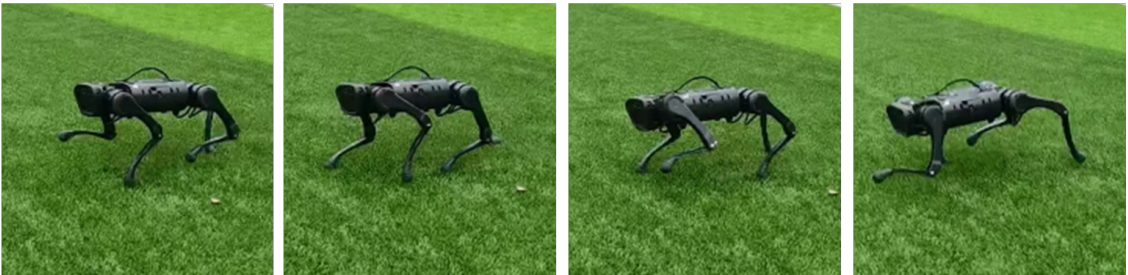

**Fig. 2. Foot trajectories at low and high speeds.**

**A**. Uneven road

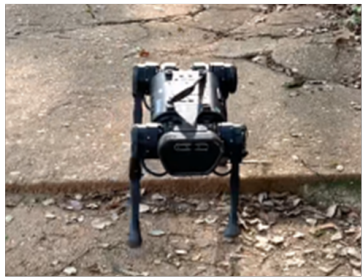

**B**. Gravel

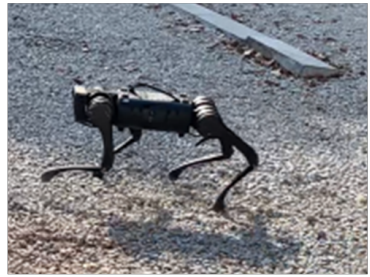

**C**. High grass

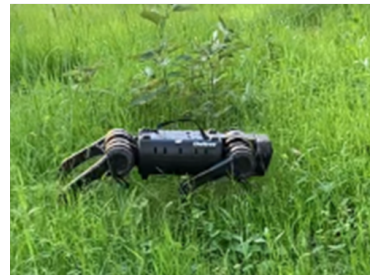

**D**. Mud and leaves

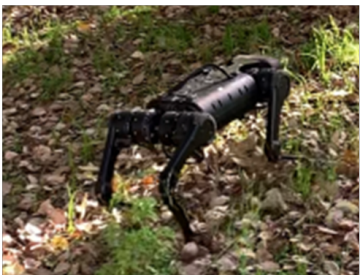

**E**. Smooth ceramic tiles

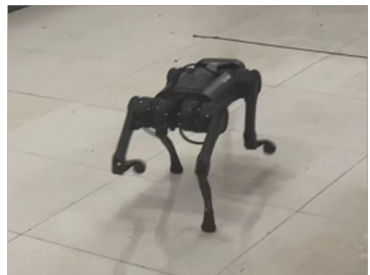

**F**. Entertainment

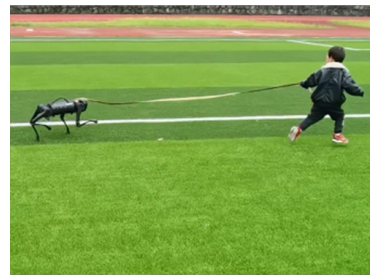

**Fig. 3. The controller can handle many complex terrains and support our entertainment mode.**

Movie S3 and Fig. 3F show an entertainment mode. In this mode, there is no remote-control command, and people can control the locomotion direction and speed of the robot through a rope. This way of directly interacting with the legged robot may bring some happiness to people, which is conducive to legged robots entering the public life. The principle used in this mode is that when the robot body is subjected to external force, the robot feels the speed change along the rope direction, the command part in the policy network input will be automatically adjusted, so as to follow the guide direction of the rope. The command speed will decay with time, and the locomotion speed will gradually decrease to 0 when the robot body is not subjected to external force. This entertainment mode is based on the strong anti-disturbance capability of our controller, thus the robot can quickly respond to the continuous external disturbance to maintain the stable movement of its body.

**Indoor experiments**

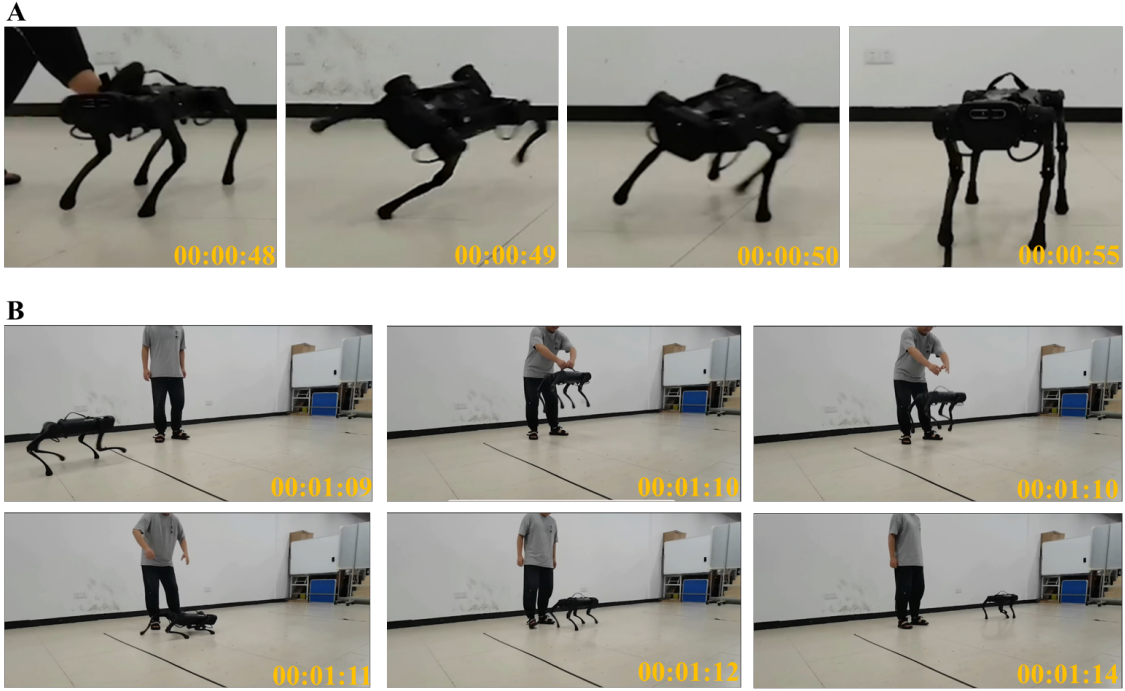

**Fig. 4. Test for anti-disturbance performance.** (**A**) Performance under strong instantaneous impact on the robot body. (**B**) Performance under a special case of hanging and falling in the middle of the locomotion. The time format in the lower right corner is "Hour: Minute: Second".

Movie S4 shows three experiments to test the anti-disturbance performance of our proposed controller. In the first two experiments, the command speed is always 0, and the controller is able to maintain the balance of the robot body no matter it encounters continuous external force in different directions or instantaneous impact. As shown in Fig. 4A, even if the roll angle is tilted more than 45 degrees and the three feet are in the air, the robot is still able to quickly adjust its posture and maintain its balance. The third experiment is that the robot is suddenly lifted up for a period of time and then falls from 70cm in the air. The process is shown in Fig. 4B. During the suspension period, foot trajectories of the robot will not diverge, and the robot can quickly adjust its posture and continue to move forward stably after landing.

The current advanced traditional human-designed controllers usually use model predictive control (MPC). In movie S5, we compare the locomotion performance of the traditional controller and our controller on smooth ground. The traditional controller will lose balance and fall directly once the foot slips when the robot passes through the wet and slippery ground. In contrast, our controller can quickly adjust the cadence and footfall when the foot slips, allowing it to traverse smooth surface with stability. In addition, our training architecture does not use artificially designed structural constraints such as the foot trajectory generator. When slippage occurs, our controller behaves that every foot tries to find reasonable support, and there will be no strange phenomena such as programmed foot suspension.

In movie S6, we compare the noise generated indoors by the traditional controller and our controller. To maintain stability, an artificially designed controller has to make a quadruped robot step at a high frequency, even when the expected body velocity is 0. As a result, the robot generates a lot of unnecessary ground-contact noise. Not only does this cause additional energy loss, but it can be annoying to people in the room. Our presented controller can make the robot move more naturally. When the command speed is gradually reduced, the robot will naturally slow down the cadence, thus reduce energy consumption and the noise caused by touching the ground. This behavior is closer to a real animal and it is helpful for the robot to enter human life.

**Validation of data symmetry loss**
The performance of the trained controller is not isotropic due to the effect of the phenomenon called data symmetry loss mentioned later. The first part of movie S7 shows the results of directly using the trained controller without our proposed mirror world network. The left side of the robot is less stable than the right side. In the state of high-speed locomotion, after slipping to the left, the robot fails to regain balance quickly and falls over. This shows that although the robot has a left-right symmetrical structure, the ability of the policy network directly obtained by simulation training is not same on the left and right sides. To further illustrate the impact of data symmetry loss on controller performance, we conduct a new training on flat ground with a new random seed. After the training is completed, the controller network is directly deployed on the physical robot, and the experimental results are shown in the second part of movie S7. In the experiment, the command velocity keeps 0. External force can be applied in different directions through the rope tied to the robot body. At the beginning, the robot stands in a stationary state. First, we slowly apply a horizontal pulling force to the left of the robot body, and then apply a horizontal pulling force to the right to observe the anti-disturbance performance of the robot. It can be found that the robot can respond to left continuous external disturbance and maintain its own stability through legs motion adjustment, but cannot respond to the continuous external disturbance in the right side, which caused it to fall. These experimental results also show that the data symmetry loss brings the performance of the trained controller unbalanced.

# DISCUSSION
In the control architecture trained by reinforcement learning, some structural constraints such as a model predictive control framework and a foot trajectory generator can be defined. In this case, the controller only needs to search parameter space, so that only a small number of parameters need to be trained, and the training can be converged using less sample data. However, the performance of the controller trained in this way is closely related to the artificially defined structural framework, which cannot guarantee the full development of the hardware potential. Our work shows that a high-performance locomotion controller can be trained without any artificially designed structural constraints.

Compared with the traditional artificially designed controller, the learned controller makes robot locomotion more natural and energy-efficient. The learned controller can achieve faster speed and have better stability and adaptability. Our robot controller trained with model-free reinforcement learning does not use any artificially designed structural constraints. This shows that it is feasible to train robot locomotion from scratch without specific human knowledge. After combining with the mirror world neural network, the robot can run stably at high speed and traverse various complex terrains. Its significance lies in that it can transfer the high dynamic continuous motion learned in the simulation to the real robot, which provides the basis for the realization of other high dynamic and complex motion of the robot in the future (For example, a robot performing Chinese kung fu is a very meaningful and cool thing).

Changes in the physical properties of the ground as well as changes in geometry can be seen as unpredictable external disturbances to the robot. Especially when the robot runs at high speed, the external disturbance is a big test for the performance of the controller. The mirror world neural network plays an important role in the stability of the robot during high-speed locomotion.

Obviously, there are other forms in the loose neighborhood control architecture, and the effect of different forms on the controller solution may be an interesting study.

From the perspective of solving nonlinear systems, the phenomenon of data symmetry loss is prevalent in training phase of reinforcement learning. This phenomenon is one of the factors that causes suboptimal solutions for nonlinear systems, and attempts to mitigate this phenomenon will result in better controller solutions. Domain randomization (*35*) not only plays the role of improving robustness by covering the real model parameters, but can also mitigate the impact of data symmetry loss (mentioned later).

The limitation of our method is that the robot only obtains the trotting gait during training. How to guide the robot to obtain other gaits through richer rewards is a problem worthy of study. The controller does not use any terrain information during the simulation training process, so the robot is not able to pass through scenes such as stairs that are highly dependent on terrain information. In future research, teacher/student strategies (*30*) can be utilized to gain the ability to traverse more challenging terrain such as stairs. Combined with hierarchical reinforcement learning (*36-39*) and multi-expert systems (*40, 41*), the capabilities of robots will be further improved.

## MATERIALS AND METHODS

### Overview

From a mathematical point of view, reinforcement learning solves the following robot nonlinear system problem:

$$\text{objective: } \max \mathrm{E}_\pi (\sum_{i=1}^{i=n} r_i) \qquad (1)$$

subject to: robotic dynamics, environment constraints, other unmodeled constraints

$r_i$ is the reward received from the environment at each time step, and $n$ is the maximum duration of an episode, $\pi$ is the action policy. Our goal is to train a robust and stable controller that enables the robot to run fast and stably and traverse various complex terrains. The controller tracks the desired body linear velocity and angular speed.

An overview of our method is shown in Fig. 5. The reinforcement learning algorithm is used to train a policy neural network in a simulation environment, as shown in Fig. 5A. The policy neural network outputs action $a_t$ according to the robot state $s_t$.

Fig. 5B is a schematic diagram of the control structure. Here we build a drive mapping. The output of the policy network does not directly act on the robot but is part of the input of the drive mapping. The drive mapping generates joint torques based on joint velocities and policy network action to drive the robot locomotion.

In the training process, three different fixed terrains are used, as shown in Fig. 5C, which are flat ground, continuously changing slopes and irregular steps. After the robot reaches a certain reward on one terrain, it switches to the next terrain in sequence.

Due to the data symmetry loss mentioned later, the trained policy neural network cannot meet the requirements of high-speed and stable motion of the physical robot. We propose to use a mirror world neural network to overcome this problem. As shown in Fig. 5D, an impact test training field is constructed in the simulation. The trained policy network is put into the training field for testing to determine the robot's advantage area, then generate a mirror world policy network according to the advantage area. Fig. 5E shows the final controller network architecture, which consists of the trained policy network, mirror world network and network selector. The controller network can be directly deployed on the physical robot.

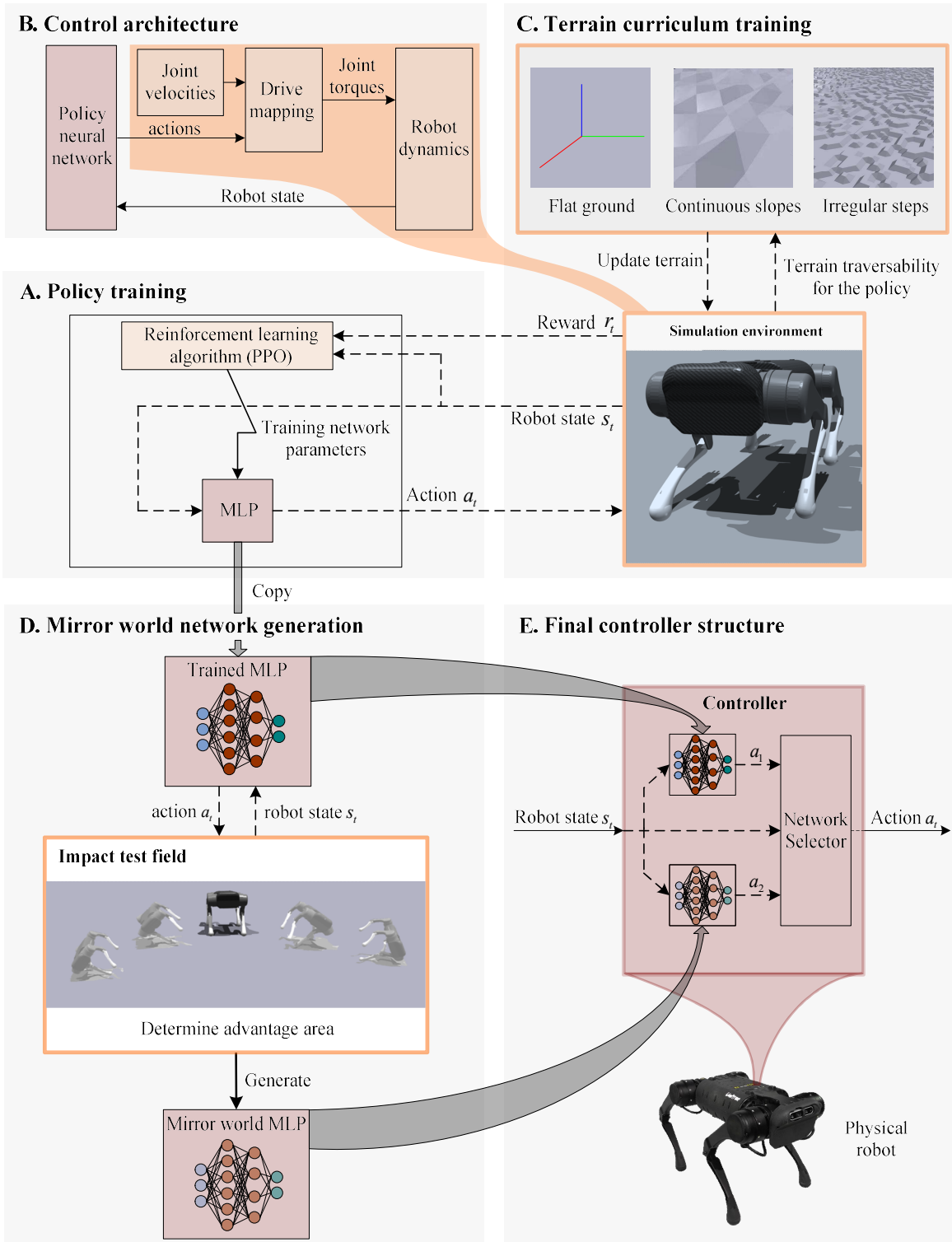

**Fig. 5. An overview of creating a locomotion controller.** (**A**) Training process. The robot continuously interacts with the simulation environment, and the reinforcement learning algorithm adjusts the network parameters in the direction of maximizing expected rewards. (**B**) Drive mapping converts joint velocities and policy network outputs into joint torques to drive the robot. (**C**) Three fixed terrains are used in the terrain curriculum training. (**D**) After the policy network is trained, it is sent to the impact test training field to determine the robot's advantage locomotion area, and then the mirror world network can be generated. (**E**) The final controller network architecture consists of the trained policy network, the mirror world network and a network selector. The controller network is directly deployed on the physical robot.

Manuscript Template    Page **10** of **23**

## Policy training

From a mathematical point of view, reinforcement learning is essentially a nonlinear solver with the basic assumption that the goal can be described by maximizing the expected reward. The training process of reinforcement learning is often modeled as a Markov decision process (MDP), which is described by a state space $S$, an action space $A$, a scalar reward function $R(s_t, a_t, s_{t+1})$, and a transition probability density $P(s_{t+1} | s_t, a_t)$. The goal is to find an optimal policy $\pi^*(a_t | s_t)$ that maximizes the expected reward. Fig. 5A shows the training process. According to the robot state $s_t$ at each time step, the policy network $\pi(a_t | s_t)$ outputs the action $a_t$, and then obtains the corresponding reward $r_t$ and the next state $s_{t+1}$ from the environment. Our reward details are shown in S2.

We use the model-free reinforcement learning approach without any artificially designed structural constraints. The robot state $s_t$ is a 48-dimensional vector, including the body velocity, the body angular speed, the attitude vector, the command velocity, joint angles, joint angular speeds and the policy network output at the last moment (Table S1). The policy network is a fully connected neural network with three hidden layers, and the output is a 12-dimensional vector (Details are listed in table S2). Domain randomization is used in training and Gauss noise is added to the action output to make the controller more robust. The policy network is trained using the PPO reinforcement learning algorithm (*18*), and the hyperparameters are listed in table S3. The training is performed on a laptop with 8 CPU cores (Intel® Core™ i7-7700HQ) and a graphics card (NVIDIA GeForce GTX1070) using the Isaac Gym simulation environment (*42*), and 4096 quadruped robots are trained in parallel (*43*). The simulation time step is 0.005s, and the maximum duration of an episode is 15s. The linear velocity command range in the training is from -2m/s to 2m/s, and the angular speed command range is from -3.14rad/s to 3.14rad/s. The control frequency is 50Hz.

## Loose neighborhood control architecture

Some previous studies use human prior knowledge to construct structured constraints. Only a small number of parameters need to be learned with the constraints, thus learning can converge quickly. However, the controller performance is limited by the artificially designed structural constraints.

Fig. 5B shows the process of converting from the output of the policy network to the robot joint torques. Unlike the previous study (*30*), no foot trajectory generator is used here, and there are no artificially designed structural constraints. The main part of the figure is the drive mapping, which accepts the output of the policy network and the joint velocities as inputs. The specific mathematical expressions are listed in S3. It may be noted that while the drive mapping looks like joint position control, it is not in fact. The actual positions of joints are weakly correlated with the output of the policy network.

The main reason for using this mapping instead of outputting joint torques directly from the policy network is that we find that learning directly in the torque space would slow the robot's learning rate and make it difficult to learn robust results. Using the drive mapping proposed here makes the joint motion space increments to be constrained within a broad neighborhood of the policy network output, which enables improved learning rates, and find the solution to the control problem while minimizing the structural constraints imposed by human prior knowledge. Since the high frequency output torque is not as easily generated as the position control, it is easier to transfer the simulation results to a physical robot.

**Terrain curriculum training**

Fig. 5C shows three terrains used in the training process, namely flat ground, continuous slopes terrain (slopes are distributed randomly between 0 degrees and 15 degrees) and irregular steps terrain (The height is 5cm). After the robot reaches a stable reward on each terrain, it sequentially transitions to the next terrain for training. Simply training on these three simple terrains can make the robot traverse terrains never seen before such as smooth surfaces, tiled floors, city roads, gravel, grass and muddy ground.

**Data Symmetry Loss Problem**

In the creatures in nature, structural symmetry is widespread. From little mice to massive elephants, dolphins in the sea, birds in the sky, including humans, are bilaterally symmetrical in appearance and structure. Functionally, however, there can be significant differences between the left side and right side of the structural symmetry. A child who is right-handed from an early age tend to perform better with his right hand than the left hand at using tableware, writing and drawing. A left-footed professional football player will usually have better ball control, passing accuracy, shooting power and accuracy with the left foot than with the right foot. The dominant side has the advantage over the side lack of training due to prolonged training. When faced with unfamiliar environments or time-sensitive tasks, people are usually more inclined to use the dominant side to take on important roles, which further leads to "the strong stronger", resulting in the increasingly pronounced asymmetry of abilities on the symmetrical structure.

This ability asymmetry also exists in the training process of reinforcement learning. Taking a quadruped robot learning omnidirectional motion as an example. The simulation training process is usually parallelized to shorten the training time (*29*). Assume that 10,000 robots are trained in parallel to collect enough training samples. During the learning process, the reset condition of each episode is set as the robot body touching the ground or reaching the maximum length of an episode. Each time the reset condition is triggered, the expected speed of the robot is given at random. At the beginning of training, the motion of the robot is determined by randomly initialized parameters of the policy network. Ideally, 5000 robots have the desired velocity direction to the right ($^B v_y < 0$), and 5000 robots have the desired velocity direction to the left ($^B v_y > 0$). Suppose the robots first learn to move to the right to survive longer without falling, thus not triggering the reset condition. Then the robots moving to the left will fall and be reset before the robots moving to the right. After reset, it will be divided into 2500 robots with a desired speed to the left and 2500 robots with a desired speed to the right. As the training continues, the 2500 robots moving to the left are reset to 1250 robots moving to the left and 1250 to the right, and so on for the same reason. We call this phenomenon of samples asymmetry in the training process, although it has structural symmetry, as data symmetry loss.

In the training process, once the robot has the advantage of right movement, it will receive more training samples in the right direction, and the number of training samples in the left direction will be squeezed. This in turn makes the right side more dominant after the robot is trained more on the right side. The final result of training is that the ability of the robot to move to the right is stronger than that of the left.

Fig. 6A shows the statistical results in the training process on flat terrain. The vertical axes are respectively the ratio of the robot's expected movement speed to the left and the expected movement speed to the right and the training reward. One Epoch on the horizontal axis represents 98,304 time steps. The trainings are carried out five times in total, and different random seeds are selected for each training and other parameters are kept unchanged. The results verify the asymmetry of the data samples during the training process. Fig. 6B are the experimental results after additionally adopting the domain randomization method on the basis of the training parameters setting in Fig. 6A. The results show that domain randomization has a certain mitigation effect on the data symmetry loss. Movie S7 shows the unbalanced locomotion ability.

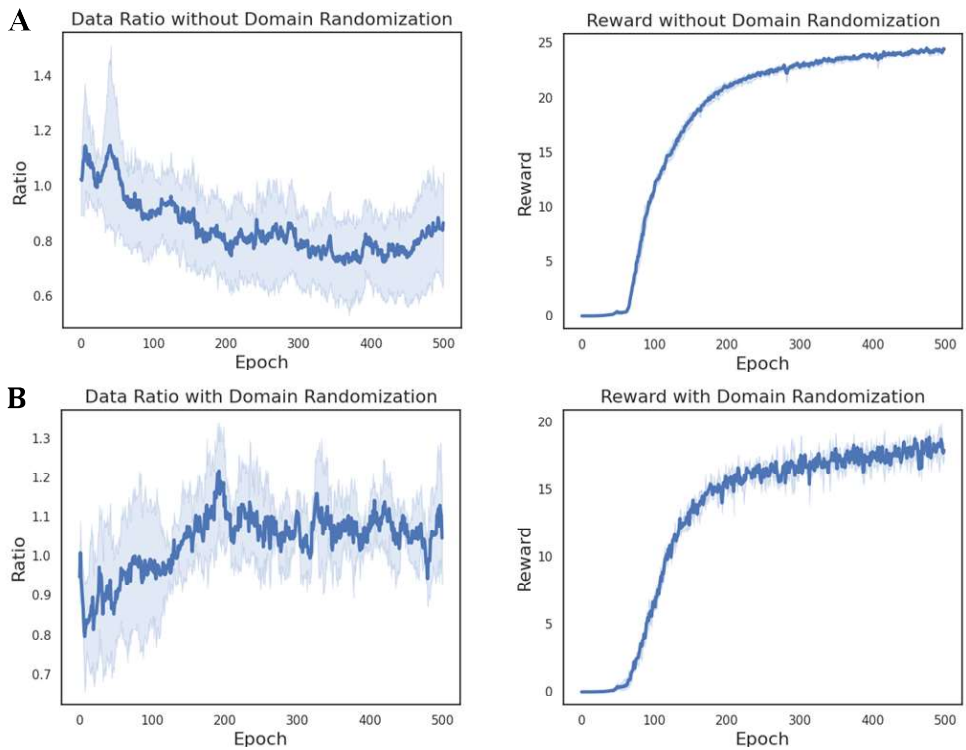

**Fig. 6. Unbalanced data samples.** (**A**) The ratio of left data samples to right data samples significantly deviates from the unit value 1.0 as the training goes on. (**B**) The ratio of left data samples to right data samples with domain randomization. Shaded areas denote 95% CIs.

We think that the reasons for the phenomenon of data symmetry loss are mainly the following aspects. First of all, in terms of mathematical nature, reinforcement learning is a nonlinear solver that continuously interacts with the environment and solves complex nonlinear system problems through sampling. However, finding the optimal solution for a general nonlinear system is still an open problem. The current reinforcement learning algorithm cannot guarantee to find the global optimal solution, and generally can only converge to the local extremum point or the saddle point and find the sub-optimal solution of the nonlinear problem. Secondly, during the training process, random numbers are generated by pseudo-random number generators, not real random numbers. These will make it impossible to keep the performance of the policy network consistent on both sides of the structure of bilateral symmetry during the training process. Once locomotion advantage on a certain side appears in the training process, there will be a phenomenon of "the strong stronger", which makes it difficult to collect random and balanced data. Unbalanced training leads to inconsistent performance of the policy network on the left and right sides of the symmetrical structure, and as the robot locomotion speed increases, the performance difference between the both sides is more obvious. The data symmetry loss puts the robot training into the case of sample preference, and the result is that the trained policy network makes the robot to exhibit advantage and weakness regions in terms of locomotion performance.

**Mirror world neural network**

Due to the data symmetry loss, the performance of the learned controller is not same on the different areas of a quadruped robot with bilateral symmetrical structure. Once the robot state falls into the weakness area, it is prone to fall, as shown in movie S7. Here, we propose a mirror world neural network to overcome the controller performance problem brought by the data symmetry loss.

Fig. 7 illustrates the basic idea of a mirror world network. Assuming that the trained network behaves as a left-side advantage area in the real world, as shown in Fig. 7A, the red colored area represents the advantage area. Just like a football player deftly plays the ball with his left foot in front of a mirror, but the player in the mirror world plays the ball with his right foot.

In Fig. 7B, from the robot's point of view, it knows that the controller it has learned performs well when it moves to the left, but does not perform well when it moves to the right. Then when its state is moving to the right, it is found that the self in the mirror world is actually moving to the left, so the trained policy network can be applied to the self in the mirror world to obtain good locomotion ability, and then the locomotion is mirrored back to the real world to obtain the real-world motion strategy, as shown in Fig. 7C, at which time the robot obtains the right advantage area (green colored part) from the mirrored world. We call the neural network that computes the output of the policy network in the mirror world and then mirrors the output back to the real world as the mirror world network. Fig. 8. illustrates the composition of the mirror world neural network. The state mirror layer and action mirror layer are both diagonal matrices whose elements is 1.0 or -

1.0.

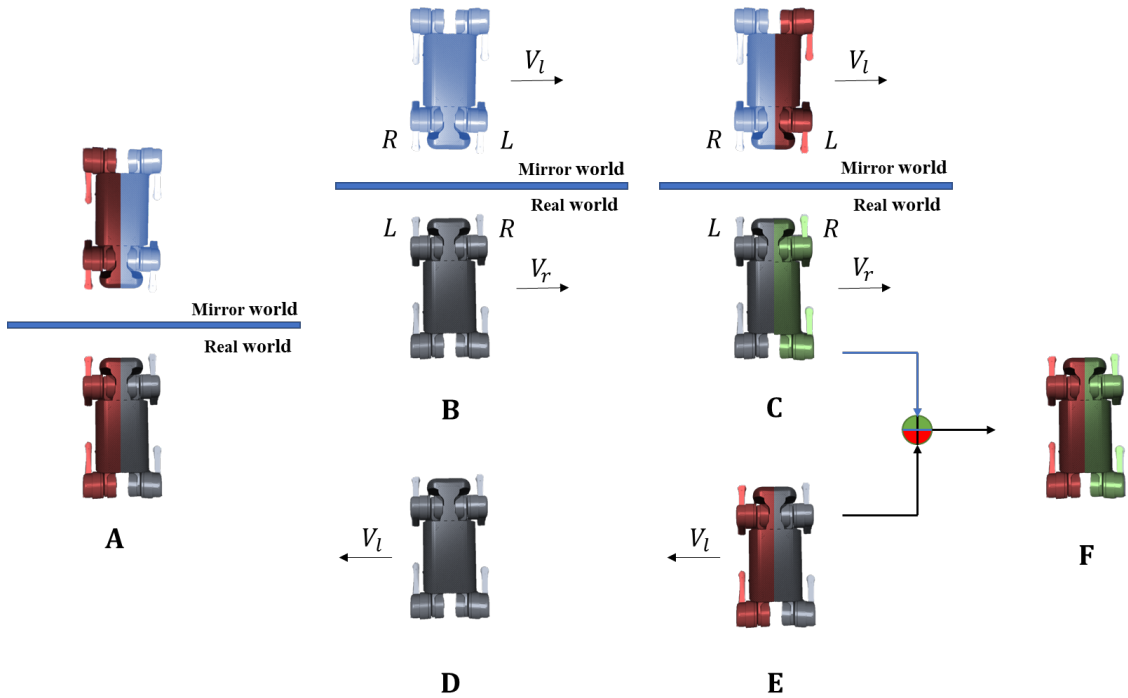

**Fig. 7. Complementary strengths in the real world and the mirror world.** (**A**) Real-world advantage in the mirror world. (**B**) Real-world rightward velocity appears as leftward velocity in the mirror world. (**C**) The advantage area on the left side of the mirror world can be transformed to the right side of the real world. (**D**) Real-world movement to the left. (**E**) The robot has a left advantage area in the real world. (**F**) Combining the real world and the advantaged area obtained from the mirror world, the robot has a bilateral advantaged area.

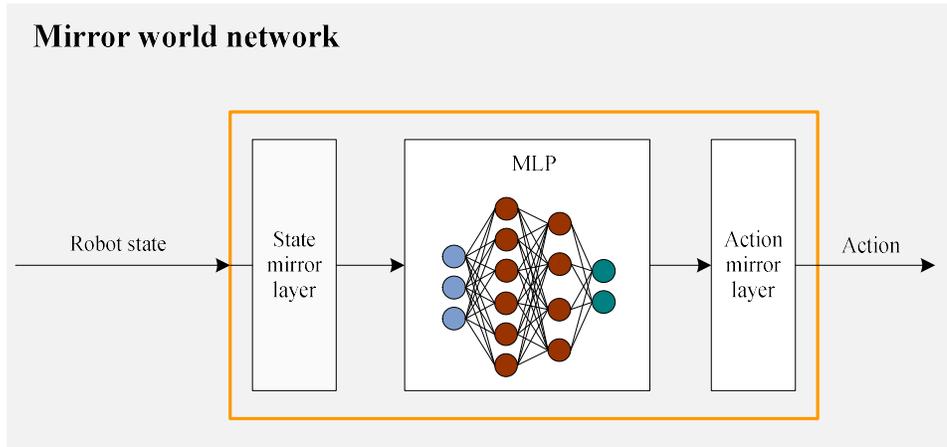

**Fig. 8. Composition of the mirror world network.**

Since the trained policy network exhibits left-side dominance, when the robot moves to the left (Fig. 7D), the policy network can be directly applied for locomotion control (Fig. 7E). By combining the trained controller and the mirror world network, a controller with bilateral advantages can be obtained, as shown in Fig. 7F.

Fig. 5E illustrate the composition of the controller network that is finally used on the physical robot. The policy network trained by reinforcement learning determines the advantage area through the impact test training field, and then the corresponding mirror world network is generated. The network output selector selects the network according to the robot state. Our controller uses whether the roll angle is greater than 0 as the network switching condition, and combines hysteresis to prevent network frequency switching. Movie S1~S5 show that our controller composed of the trained controller and the mirror world network has excellent locomotion performance and anti-disturbance ability.

**Simulation to real**

The joints of the UnitreeA1 quadruped robot are driven by DC motors. Our method mainly focuses on making the simulation training results as easy as possible to transfer to the real robot without modeling the actuators.

Transfer from simulation to real world is an important issue. A rigid body simulation environment can only simulate the properties of real-world objects to a certain extent. The calculation results of the physics engine are affected by the calculation method and accuracy, and there is also a certain deviation from the laws of the real physical world. Furthermore, differences between the URDF model and the physical model are unavoidable, and joint actuation forces in the simulation are instantaneous, while real-world actuators have response time and steady-state errors. The influence of these differences on the behavior of the physical robot will become more obvious with the increase of the joint motion speed of the robot, which directly affects the robot performance in high-speed motion.

The solution of a policy network obtained by training in the simulation environment is a continuous function, and there are infinitely many solutions with similar performance near this function. An obvious conclusion is that smooth solutions are easier to transfer to a physical robot. In fact, the driving torques in the simulation are instantaneous and ideal, while the program delay, the delay caused by actuator mechanical structure and the control loop in the real world are inevitable. If the output of the policy network trained in the simulation contains a lot of high frequency components, it will be difficult for the physical actuator to track the network output. In contrast, even if the output of the learned controller contains a large number of high-frequency components, the robot can achieve the predetermined goal in the simulation environment. But to make it easier to transfer the results to the real robot, a solution network that contains as less high-frequency components as possible is a better choice.

Our loose neighborhood control architecture takes into account both training rate and sim-to-real. It can constrain the robot's position configuration within a broad region to improve learning speed, without the high frequency component that is easy to occur in pure joint position control. By combining reward penalty, loose neighborhood control, domain randomization, and adding noise at the output of the policy network, the trained controller can be directly deployed on UnitreeA1.

**Supplementary Materials**
Section S1. Nomenclature
Section S2. Reward function for policy training
Section S3. Drive mapping
Section S4. Algorithm in the impact test training field
Table S1. State representation of the controller input.
Table S2. Policy neural network architecture.
Table S3. PPO hyperparameters for policy training.
Movie S1 (.mp4 format). Stability at high speed.
Movie S2 (.mp4 format). Passability test for complex terrains.
Movie S3 (.mp4 format). Entertainment mode.
Movie S4 (.mp4 format). Stability under external disturbance.
Movie S5 (.mp4 format). Comparison of the robustness by different controllers.
Movie S6 (.mp4 format). Comparison of indoor noise generated by different controllers.
Movie S7 (.mp4 format). Unbalanced locomotion ability caused by data symmetry loss.

**S1. Nomenclature**
- $v$ linear velocity
- $\omega$ angular speed
- $\theta$ joint angle
- $\tau$ joint torque
- $e_g$ unit vector along the direction of gravity
- $^B(\cdot)$ vector in body frame
- $^w(\cdot)$ vector in world frame
- $^B_w T$ direct cosine matrix from world frame to body frame
- $\|\cdot\|_2$ $l_2$ norm

**S2. Reward function for policy training**
During training, the goal of reinforcement learning is to maximize the expected reward. The reward function is defined as $r = r_{aim} + r_{penalty}$, where $r_{aim} = r_{lv} + 0.5 r_{az}$ is the tracking reward, $r_{penalty} = 3 r_{lvp} + 0.05 r_{azp} + r_g + 5e^{-4} r_\tau + 0.25 r_{collide} + 0.1 r_{ar}$ is the penalty reward. They are defined as follows.

**Tracking reward**
$r_{lv}$ is the linear velocity tracking reward in the body frame. The closer the body linear velocity in horizontal plane $v_{xy}$ and the command linear velocity $v_{cxy}$ are, the higher the reward.

$$r_{lv} = \exp(-3 \|^B v_{xy} - v_{cxy}\|_2) \quad (2)$$

$r_{az}$ is the tracking reward along z axis in the body frame. The closer the body angular speed $\omega_z$ and the command $\omega_{cz}$, the higher the reward.

$$r_{az} = \exp(-3(^B\omega_z - \omega_{cz})^2) \quad (3)$$

**Penalty reward**
$r_{lvp}$ is the linear velocity penalty reward along z axis in the body frame. The robot body is encouraged to maintain a stable position in the vertical direction.

$$r_{lvp} = -v_z^2 \tag{4}$$

$r_{azp}$ is the angular speed penalty in the body frame. It encourages the robot to reduce amplitude of the roll and pitch angular speeds.

$$r_{azp} = -(\omega_x^2 + \omega_y^2) \tag{5}$$

$r_g$ is the body attitude penalty. It encourages the roll and pitch amplitudes of the robot to be as small as possible.

$$r_g = -\left\|({}_w^B Te_g)_{xy}\right\|_2 \tag{6}$$

$r_\tau$ is the joint torque penalty. It encourages the robot to consume as little energy as possible to complete the task.

$$r_\tau = -\sum_i |\tau_i| \tag{7}$$

$r_{collide}$ is the rigid body collision penalty. $r_{collide} = -1$ if the knee touches the ground, else $r_{collide} = 0$.

$r_{ar}$ is the motion smoothness penalty. The policy neural network is encouraged to have adjacent outputs $a_t$ and $a_{t+1}$ as close as possible.

$$r_{ar} = -\|a_{t+1} - a_t\|_2 \tag{8}$$

**S3. Drive mapping**
The mathematical description of the drive mapping is

$$\tau_t = k_p(a_t + \theta_0 - \theta_t) - k_d(\dot{\theta}_t) \tag{9}$$

$a_t$ is the policy neural network output vector at time step $t$. $\theta_0$ is the initial joint angle vector. $\theta_t$ is the joint angle vector at time step $t$. $\dot{\theta}_t$ is the joint angular speed vector at time step $t$. $k_p$ and $k_d$ are both scalars.

**S4. Algorithm in the impact test training field**
1. Initialize the robot state, the expected linear velocity and the expected angular velocity are set both 0.
    testcounter = 0
    totoaltestnumber = 30
    falltoleft = 0
    falltoright = 0
    leftadvantage = False
    $V_{max} = 1$
2. while testcounter < totaltestnumber :
    if testcounter % 2 == 0 :
        Every once in a while $t$, push the robot to the right to speed $V_y \in (-V_{max}, 0)$
        If the robot falls, then falltoright = falltoright + 1, reset
    else :
        Every once in a while $t$, push the robot to the left to speed $V_y \in (0, V_{max})$
        If the robot falls, then falltoleft = falltoleft + 1, reset
    testcounter = testcounter + 1

3. if falltoright > falltoleft :
       return leftadvantage = True
   else if falltoright < falltoleft :
       return leftadvantage = False
   else if falltoright == falltoleft and falltoright != 0 :
       $V_{max} = 0.5 V_{max}$
   else :
       $V_{max} = 1.5 V_{max}$
4. Back to step 1

| Variables | Dimension |
|---|---|
| Body linear velocity | 3 |
| Body angular speed | 3 |
| Command | 3 |
| Attitude vector | 3 |
| Joint position | 12 |
| Joint velocity | 12 |
| Last action | 12 |

Table S1: **State representation of the controller input.**

| Layer | Type |
|---|---|
| 1 | Input [Identity(48)] |
| 2 | ELU(512) |
| 3 | ELU (256) |
| 4 | ELU (128) |
| 5 | Output [Linear(12)] |

Table S2: **Policy neural network architecture.** ELU is short for Exponential Linear Unit.

| Parameters | Value |
|---|---|
| discount factor | 0.99 |
| clip value ($\varepsilon$) | 0.2 |
| KL-d threshold | 0.008 |
| learning rate* | 3e-4 |
| batch size | 98304 |

Table S3: **PPO hyperparameters for policy training.** The Adam (*44*) optimizer is used. Learning rate with * is a dynamic learning rate, it decays linearly with training epochs and the minimum value is 1e-6.

**Acknowledgments:**
    **Author contributions:**
        Conceptualization: Xu Chang
        Methodology: Xu Chang, Zhitong Zhang
        Formal analysis: Xu Chang, Zhitong Zhang, Honglei An, Hongxu Ma
        Investigation: Xu Chang, Zhitong Zhang, Honglei An
        Visualization: Xu Chang, Zhitong Zhang, Hongxu Ma
        Project administration: Hongxu Ma, Honglei An, Qing Wei
        Supervision: Xu Chang, Zhitong Zhang, Hongxu Ma
        Writing – original draft: Xu Chang
        Writing – review & editing: Zhitong Zhang, Honglei An, Qing Wei

X. C. formulated the main idea of the training and control architecture, proposed all conceptions and experiments. Z. Z. proposed a main idea in the training and performed all the experiments. X. C. and Z. Z. implemented the controller together. All authors contributed in setting up the simulation, refined ideas, contributed in the experiment design, and analyzed the data.

**Competing interests:** The authors declare that they have no competing interests.

**Data and materials availability:** All data needed to evaluate the conclusions in the paper are present in the paper or the Supplementary Materials.